\documentclass[11pt]{article}

\usepackage[preprint]{acl}
\usepackage[T1]{fontenc}
\usepackage{times}
\usepackage{latexsym}
\usepackage{amsmath}
\usepackage{amssymb}
\usepackage{booktabs}
\usepackage{xcolor}
\usepackage{colortbl}
\definecolor{cceblue}{HTML}{E3F1FA}
\usepackage{placeins}
\usepackage{float}
\usepackage{multirow}
\usepackage{graphicx}
\usepackage{pgfplots}
\usepgfplotslibrary{groupplots}
\pgfplotsset{compat=1.18}
\usepackage{microtype}
\usepackage{inconsolata}
\usepackage{listings}
\usepackage{algorithm}
\usepackage{algpseudocode}
\lstdefinestyle{cceprompt}{
  basicstyle=\ttfamily\scriptsize,
  breaklines=true,
  breakatwhitespace=false,
  columns=fullflexible,
  keepspaces=true,
  showstringspaces=false,
  frame=single,
  rulecolor=\color{black!25},
  aboveskip=5pt,
  belowskip=5pt
}

\title{Condition-Contrastive Extrapolation for On-Policy Distillation of Language Models}

\author{
  Wenxuan Luo\textsuperscript{1}\thanks{Equal contribution.} \quad
  Chenxi Li\textsuperscript{1}\footnotemark[1] \quad
  Haoyue Liu\textsuperscript{1,2}\footnotemark[1] \\
  Yongxin Guo\textsuperscript{4} \quad
  Xiaoying Tang\textsuperscript{1,2,3}\thanks{Corresponding author.} \\
  {\small\textnormal{\textsuperscript{1}School of Science and Engineering, The Chinese University of Hong Kong, Shenzhen}} \\
  {\small\textnormal{\textsuperscript{2}Shenzhen Future Network of Intelligence Institute (FNii-Shenzhen)}} \\
  {\small\textnormal{\textsuperscript{3}Guangdong Provincial Key Laboratory of Future Networks of Intelligence, CUHK(SZ)}} \\
  {\small\textnormal{\textsuperscript{4}ByteDance}} \\
  {\small\texttt{wenxuanluo@link.cuhk.edu.cn} \quad\texttt{tangxiaoying@cuhk.edu.cn}}
}

\begin{document}
\maketitle
\raggedbottom

\begin{abstract}
On-policy distillation (OPD) improves a student language model by training on its own generated trajectories with supervision from a stronger teacher. However, when conditioned on student-generated prefixes, the teacher can favor invalid continuations that violate the question's premises, causing OPD to encourage erroneous reasoning steps. As a remedy, we introduce \textbf{Condition-Contrastive Extrapolated On-Policy Distillation (CCE-OPD)}, which makes teacher supervision sensitive to question conditions. CCE-OPD incorporates (1) a fixed-prefix comparison in which the same frozen teacher evaluates the complete question and a counterpart with a selected condition masked, and (2) a bounded log-probability contrast that reweights the teacher distribution into an auxiliary distillation target. On 200 diagnostic reasoning states, condition contrast reduces the rate of preferring invalid continuations from 27.0\% to 16.0\%. Extensive experiments across six mathematical benchmarks show that CCE-OPD improves avg@8 over OPD by an average of 1.60 and 1.95 percentage points when distilling Qwen3-8B into 0.6B and 1.7B students, respectively. It also achieves the strongest results on four transfer benchmarks.
\end{abstract}

\begin{figure*}[!t]
\centering
\includegraphics[width=\textwidth]{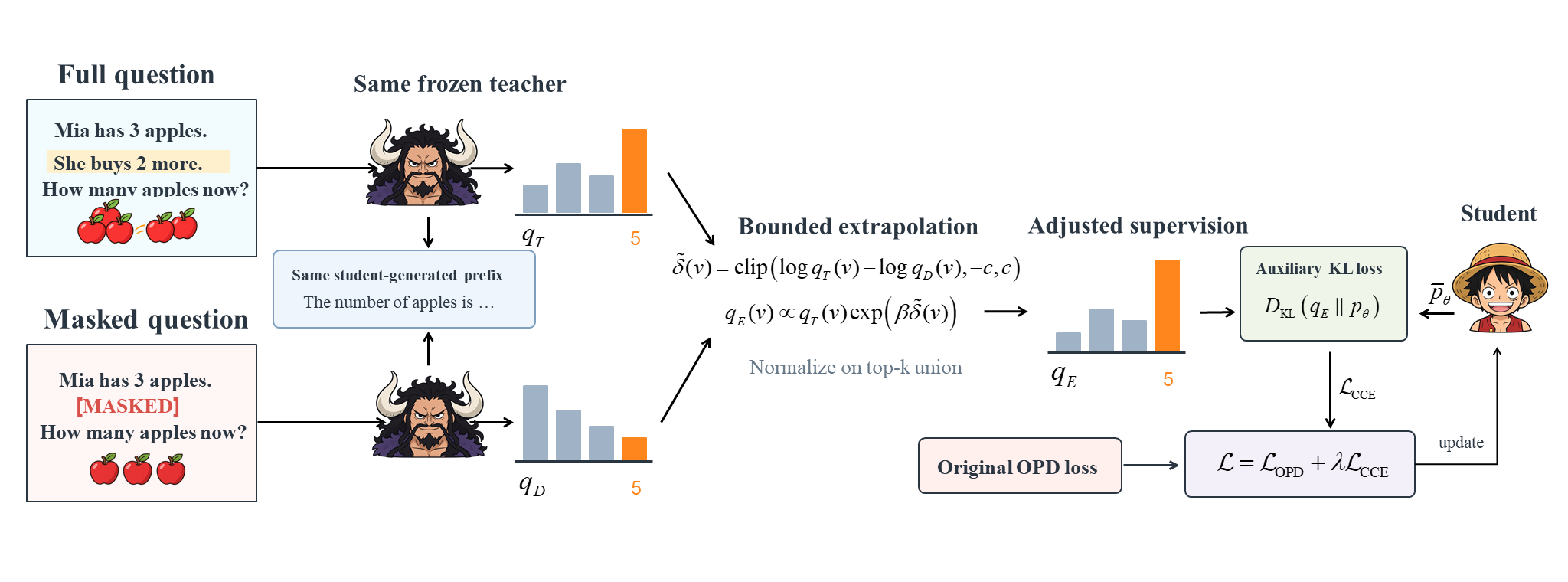}
\caption{Overview of CCE-OPD. The same frozen teacher evaluates a shared student prefix under the complete question and a version with a selected condition masked. The clipped condition contrast reweights the full-teacher distribution to form the auxiliary target $q_E$. The student is trained with the resulting auxiliary KL loss together with the original OPD objective.}
\label{fig:cce-method}
\end{figure*}

\section{Introduction}

Knowledge distillation \citep{hinton2015distilling} transfers the predictive behavior of a larger teacher to a smaller student, offering a practical route to reducing the cost of deploying capable language models. For autoregressive generation, supervision can take the form of teacher-produced responses or token distributions that convey preferences among alternative continuations. Training on teacher-generated trajectories, however, exposes the student to prefixes that can differ from those it produces at inference time \citep{agarwal2023gkd}. This mismatch is particularly relevant to multi-step reasoning, where each generated step determines the context for subsequent decisions.

On-policy distillation (OPD) addresses this mismatch by collecting trajectories from the student and querying the teacher at the resulting prefixes \citep{agarwal2023gkd}. The teacher thus provides feedback at states the student actually visits, including those arising from its own mistakes. Sampled reverse-KL objectives turn teacher--student log-probability differences into dense token-level learning signals \citep{li2026rethinking}. Recent approaches refine this supervision through reliable regions of teacher--student agreement \citep{xing2026trust} and additional distributional matching at high-entropy positions \citep{jin2026entropy}. These developments make OPD a useful framework for transferring reasoning capabilities while accounting for the student's evolving generation behavior.

However, these approaches improve how teacher feedback is applied on student-generated trajectories, but they do not by themselves ensure that a preferred continuation satisfies the question's conditions. We refer to continuations whose reasoning conflicts with the question's stated conditions as \textbf{condition-violating continuations (CVC)}. Our diagnostics show that, conditioned on a student-generated prefix, the teacher can favor CVC that follow an erroneous reasoning direction. For example, suppose a question specifies $x>0$ and $x^2=9$. After a student prefix such as ``Taking the negative root, $x=$'', continuing with ``$-3$'' follows the prefix's reasoning direction but violates the question's positivity constraint. We find that OPD can assign a positive local advantage to a sampled token that begins a CVC, producing a learning signal that favors that continuation. To refine this supervision, we seek to distinguish continuations supported by the question's conditions from those that merely follow the student's reasoning direction. We therefore keep the student prefix fixed and compare the teacher's predictions with and without a selected condition. The resulting contrast provides an additional signal for reweighting the distillation target toward valid continuations.

We propose Condition-Contrastive Extrapolated On-Policy Distillation (CCE-OPD) to refine teacher supervision on student trajectories (Figure~\ref{fig:cce-method}). At the same student prefix, a frozen teacher scores the complete question and a version with a selected condition masked. Their clipped condition contrast reweights the teacher distribution into an auxiliary target that complements sampled-token OPD.

Our contributions are threefold:
\begin{itemize}
\item \textbf{Diagnosing supervision errors.} We identify teacher preferences for CVC on student trajectories and use fixed-prefix comparisons to examine how question conditions affect those preferences.
\item \textbf{Refining continuation preferences.} We introduce CCE-OPD, which uses clipped condition contrast to reweight an auxiliary distillation target.
\item \textbf{Evaluating effectiveness and design choices.} Across two student scales, CCE-OPD achieves the best or tied-best avg@8 on six mathematical benchmarks. Input-construction, extrapolation-strength, and entropy-gate ablations assess its design choices, and four transfer tasks evaluate generalization beyond mathematics.
\end{itemize}

\section{Preliminaries}

\paragraph{Student trajectories and teacher feedback.}
Let $x$ be a question and $p_\theta$ an autoregressive student over vocabulary $\mathcal V$. A rollout policy $p_{\mathrm{old}}$, held fixed during each update, generates a response $\hat y=(\hat y_1,\ldots,\hat y_T)$. At position $t$, the state is $s_t=(x,\hat y_{<t})$, containing the complete question and the student's preceding tokens. A frozen teacher $q_T(\cdot\mid s_t)$ evaluates continuations at the same state. Both models therefore condition on the student's trajectory, while $p_\theta$ is optimized using the collected responses \citep{agarwal2023gkd}.

\paragraph{Sampled-token on-policy distillation.}
At a fixed state $s$, reverse-KL distillation measures the discrepancy from the student distribution to the teacher distribution:
\begin{equation}
D_{\mathrm{KL}}(p_\theta\|q_T\mid s)
=\sum_{v\in\mathcal V}p_\theta(v\mid s)
\log\frac{p_\theta(v\mid s)}{q_T(v\mid s)}.
\label{eq:reverse-kl-prelim}
\end{equation}
Sampled-token OPD uses teacher--student log-probability differences on generated tokens to provide local learning signals. For a token sampled from $p_{\mathrm{old}}$, the advantage is \citep{li2026rethinking}
\begin{equation}
A_t=\log q_T(\hat y_t\mid s_t)-\log p_{\mathrm{old}}(\hat y_t\mid s_t).
\label{eq:opd-advantage}
\end{equation}
The teacher scores and rollout probabilities are held fixed during the update. The current student is optimized through a clipped policy ratio $r_t(\theta)=p_\theta(\hat y_t\mid s_t)/p_{\mathrm{old}}(\hat y_t\mid s_t)$; we denote this sampled reverse-KL training objective by $\mathcal L_{\mathrm{OPD}}$. Before clipping becomes active, a positive $A_t$ encourages the sampled token and a negative $A_t$ discourages it. This sign reflects the teacher's relative preference at the given prefix, which is the local supervision signal examined in Section~\ref{sec:condition-diagnosis}.

\paragraph{Distributional supervision.}
A distributional target $q_\star$ provides feedback on alternative next tokens as well as the sampled token. A forward-KL auxiliary objective at state $s$ takes the form
\begin{equation}
D_{\mathrm{KL}}(q_\star\|p_\theta\mid s)
=\sum_{v\in\mathcal V}q_\star(v\mid s)
\log\frac{q_\star(v\mid s)}{p_\theta(v\mid s)}.
\label{eq:forward-kl-prelim}
\end{equation}
For a fixed target, minimizing this objective matches the student's relative probabilities to the target weights. CCE-OPD constructs such an auxiliary target from two teacher predictions, normalizing the target on a restricted candidate support while retaining full-vocabulary student probabilities, as defined in Section~\ref{sec:cce-method}. In the following diagnostics, $p_S$ denotes the frozen rollout student and thus plays the role of $p_{\mathrm{old}}$ in Equation~\ref{eq:opd-advantage}.

\section{Diagnosing Teacher Continuation Preferences}
\label{sec:condition-diagnosis}

We first examine teacher feedback at the decision immediately preceding a verified mathematical error (Table~\ref{tab:error-signals}), then evaluate teacher preferences over valid continuations and CVC across a larger diagnostic set (Table~\ref{tab:condition-ranking}).

\subsection{A Local Error Before It Is Committed}
\label{sec:error-signal}

The question gives $3^{400}\equiv1\pmod{1000}$ and asks for the last three digits of $3^{12000}$. After restating the condition, the student writes: ``This tells us that the order of 3 modulo 1000 is 400.'' This is the first false assertion: the congruence implies that the order divides 400, whereas its actual value is 100. Appendix~\ref{app:error-context} gives the mathematical verification and the exact decision context.

We truncate the student response immediately before the first digit of 400. The retained prefix ends with ``the order of 3 modulo 1000 is **''. The token \texttt{4} begins the CVC; the alternative \texttt{1} begins the valid assertion with value 100. The teacher receives this identical prefix under the complete question ($q_T$) and a question with the given congruence masked ($q_D$). We also form the auxiliary target $q_E$ using the bounded extrapolation in Section~\ref{sec:cce-method}.

In Table~\ref{tab:error-signals}, the full teacher assigns the token beginning the CVC more probability than the student does, giving $A_t=\log(0.597318/0.518624)\approx+0.141$. The resulting local OPD signal favors the CVC. Masking the condition raises this token's probability to 89.61\% and lowers the alternative to 1.03\%. The valid candidate consequently has a larger condition contrast $\delta_t$. Incorporating that contrast reverses their ordering: $q_E$ assigns 48.96\% to \texttt{1} and 43.80\% to \texttt{4}. 

This example shows that condition contrast can redirect the auxiliary target toward a valid continuation despite the full teacher's preference for CVC.

\begin{table}[!t]
\centering
\footnotesize
\setlength{\tabcolsep}{3pt}
\begin{tabular}{@{}lrrr@{}}
\toprule
Source & \texttt{4} (CVC) & \texttt{1} (valid) & Other \\
\midrule
Student $p_S$ & 51.8624 & 30.2846 & 17.8530 \\
Full teacher $q_T$ & 59.7318 & 31.1267 & 9.1415 \\
Masked teacher $q_D$ & 89.6143 & 1.0278 & 9.3579 \\
\rowcolor{cceblue} Target $q_E$ & 43.7999 & \textbf{48.9643} & 7.2358 \\
\bottomrule
\end{tabular}
\caption{Decision-token probabilities (\%) at the same error-before state. Labels refer to the complete assertions with values 400 and 100. The first three rows use full-vocabulary probabilities; $q_E$ is normalized on the teacher candidate union with $\beta=0.2$ and $c=5$. ``Other'' contains all remaining probability mass.}
\label{tab:error-signals}
\end{table}

\subsection{Teacher Preferences Across Reasoning States}
\label{sec:paired-diagnostic}

To extend the analysis in Section~3.1 beyond a single example, we evaluate 200 reasoning states. At each state, we compare two candidate next steps under the same question and student prefix: a valid inference and a CVC candidate. In 60 states, the student's original next step is erroneous. We score each complete candidate by averaging its token log probabilities. The CVC preference rate is the percentage of states where the CVC candidate scores higher.

We compare scores from the student and from the teacher with the complete or masked question. We also test whether adding condition contrast improves the full-teacher score. For each token, we subtract the masked-input log probability from the full-input log probability and clip the difference to $[-5,5]$. We add $0.2$ times the clipped contrast to each full-teacher token log probability, subtract the corresponding log normalizer, and average the resulting normalized log probabilities over the candidate. We repeat the same procedure with length-matched random masking as a control. Appendix~\ref{app:paired-context} provides the construction and scoring details.

As shown in Table~2, the full teacher prefers the CVC candidate in 54 of 200 states (27.00\%). Among the 60 states where the student's next step is erroneous, the teacher prefers CVC in 49 states (81.67\%), compared with 56 states (93.33\%) for the student. Thus, the teacher does not consistently favor a valid next step when the student makes an error.

Adding condition contrast lowers the CVC preference rate from 27.00\% to 16.00\% overall and from 81.67\% to 45.00\% in the error subset. By comparison, random-masking contrast lowers the overall rate only to 25.50\%. All comparisons use the same prefixes and candidate pairs. The larger improvement from condition contrast motivates its use for refining teacher supervision.

To examine how the aggregate reduction arises, we compare the two rankings state by state. Condition contrast corrects 23 full-teacher preferences for CVC and reverses one previously correct preference, yielding a net reduction of 22 states. All 24 changes occur within the 60-state subset whose naturally generated next inference is erroneous. Rankings remain unchanged in the other 140 states, including five that favor CVC under both scoring schemes. Thus, the improvement is concentrated at the student error states central to our diagnosis: condition contrast shifts the teacher's preference toward a valid alternative at the same prefix. Appendix~\ref{app:paired-context} reports the paired counts.

\begin{table}[!t]
\centering
\footnotesize
\setlength{\tabcolsep}{3pt}
\begin{tabular}{@{}lrr@{}}
\toprule
Scoring source & All 200 & Error subset (60) \\
\midrule
Student & 61 (30.50) & 56 (93.33) \\
Full teacher & 54 (27.00) & 49 (81.67) \\
Masked teacher & 64 (32.00) & 57 (95.00) \\
\rowcolor{cceblue} Teacher + condition contrast & \textbf{32 (16.00)} & \textbf{27 (45.00)} \\
Teacher + random contrast & 51 (25.50) & 46 (76.67) \\
\bottomrule
\end{tabular}
\caption{States preferring CVC: count (\%). Lower is better. The second column covers all diagnostic states; the third is the subset whose naturally generated next step is erroneous. All sources score the same candidate pairs.}
\label{tab:condition-ranking}
\end{table}

\section{Condition-Contrastive Extrapolated OPD}
\label{sec:cce-method}

Motivated by the diagnostic findings in Section~3, we introduce CCE-OPD to refine teacher continuation preferences through condition contrast. We first construct complete and masked question pairs and compare teacher predictions at the same student prefix. We then use bounded extrapolation to form the auxiliary target and apply it at high-entropy positions alongside the original OPD objective.

\subsection{Masked Inputs}

We first construct a question counterpart with a selected condition masked. For each training question $x$, a DeepSeek-Flash annotator with thinking disabled proposes a minimal condition and identifies the requested output. Exact-span checks and semantic review retain conditions whose removal changes an answer constraint or necessary solution path while preserving the request. Masking the accepted span with a fixed marker yields $x^{-}$, with masked spans denoted by \texttt{[MASKED]} in Figure~\ref{fig:cce-method}. These counterparts are fixed before training.

Of 7,498 training questions, 3,402 have accepted counterparts. They receive both OPD and the auxiliary loss of CCE-OPD; the remaining 4,096 receive OPD. The student always generates from the complete question. Construction details appear in Appendix~\ref{app:training-details}.

\subsection{Condition Contrast}

With these question pairs fixed, we compare teacher predictions at the same student prefix $\hat y_{<t}$. The frozen teacher scores both views:
\begin{equation}
\begin{aligned}
q_T(v)&=q(v\mid x,\hat y_{<t}),\\
q_D(v)&=q(v\mid x^{-},\hat y_{<t}).
\end{aligned}
\end{equation}
We define the condition contrast as
\begin{equation}
\delta_t(v)=\log q_T(v)-\log q_D(v).
\label{eq:condition-contrast}
\end{equation}
A positive contrast means that candidate $v$ receives more probability when the selected condition is present in the question. Candidates with larger clipped contrasts gain relative weight over the full-teacher distribution. Both evaluations retain the exact student prefix, including any earlier restatement of that condition.

\subsection{Bounded Extrapolation}

To convert this contrast into an auxiliary target, we reweight the full-teacher distribution with a bounded adjustment. Let $\mathcal V_t$ be the union of the top-$k$ candidates from $q_T$ and $q_D$. With $k=16$, the support has at most 32 tokens. We clip the contrast, $\tilde\delta_t(v)=\operatorname{clip}(\delta_t(v),-c,c)$, and construct
\begin{equation}
q_E(v)=\frac{q_T(v)\exp(\beta\tilde\delta_t(v))}{\sum_{u\in\mathcal V_t}q_T(u)\exp(\beta\tilde\delta_t(u))},
\label{eq:extrapolation}
\end{equation}
for $v\in\mathcal V_t$. The complete-input teacher supplies the base distribution, and $\beta$ controls the strength of the condition adjustment. We use $\beta=0.2$ and $c=5$.

To see how this target changes local preferences, consider two candidate tokens $a,b\in\mathcal V_t$ at the same student prefix. Their shared normalizer cancels, giving
\begin{equation}
\log\frac{q_E(a)}{q_E(b)}=\log\frac{q_T(a)}{q_T(b)}+\beta[\tilde\delta_t(a)-\tilde\delta_t(b)].
\label{eq:pairwise-preference}
\end{equation}
The target thus combines the teacher's original log odds with a condition-dependent adjustment. For $\beta>0$, if token $a$ begins a valid continuation and token $b$ begins a CVC, a larger clipped contrast for $a$ increases its weight relative to $b$. If the contrastive adjustment is large enough to overcome the teacher's initial preference for $b$, the target assigns higher probability to $a$ than to $b$. This relation explains how the contrast can redirect the local preference in Section~\ref{sec:error-signal}: the valid candidate gains relative support because its probability depends more strongly on the selected condition.

Without clipping, the target has the ratio form
\begin{equation}
q_E(v)\propto q_T(v)\left(\frac{q_T(v)}{q_D(v)}\right)^\beta.
\label{eq:unclipped-extrapolation}
\end{equation}
It extrapolates in log-probability space from the prediction under the masked question through the complete prediction. Clipping bounds each multiplicative adjustment to $[e^{-\beta c},e^{\beta c}]$ before normalization. At $\beta=0$, the target is the full teacher renormalized on $\mathcal V_t$, which provides the matched reweighting control in Section~\ref{sec:beta-ablation}.

\subsection{Entropy-Gated Auxiliary Supervision}

Having constructed the auxiliary target $q_E$, we next determine where to apply its supervision. High entropy indicates a more dispersed teacher distribution, with multiple continuation candidates competing for probability mass, giving condition contrast more scope to shift their relative preferences. We therefore use an entropy gate to concentrate auxiliary supervision on these positions.

Specifically, we compute the full-input teacher entropy over the complete vocabulary $\mathcal V$:
\begin{equation}
H(q_T)=-\sum_{v\in\mathcal V}q_T(v)\log q_T(v).
\end{equation}
Entropy uses natural logarithms and is computed before top-$k$ support construction and renormalization. The gate therefore evaluates the original teacher distribution $q_T$, while the auxiliary loss uses the extrapolated target $q_E$. We retain the fixed threshold $\tau=0.8$ nats from the existing auxiliary-loss training configuration and keep it unchanged across gated runs. Section~\ref{sec:entropy-ablation} assesses this selection rule by comparing it with applying auxiliary supervision at every valid position on eligible questions.

On questions with an accepted counterpart, the auxiliary loss is active at valid response positions satisfying $H(q_T)\geq\tau$. Let $m_t$ indicate positions satisfying these three conditions and $N$ be the number of all valid response tokens in the update. The target $q_E$ is normalized on $\mathcal V_t$ and assigned zero probability outside it, while the student distribution $p_{\theta,t}$ retains full-vocabulary normalization. The loss is
\begin{equation}
\mathcal L_{\mathrm{CCE}}=\frac{1}{N}\sum_t m_tD_{\mathrm{KL}}(q_E\|p_{\theta,t}),
\end{equation}
\begin{equation}
\mathcal L=\mathcal L_{\mathrm{OPD}}+\lambda\mathcal L_{\mathrm{CCE}},\qquad\lambda=1.
\label{eq:cce-objective}
\end{equation}
Sampled-token OPD supplies supervision throughout the trajectory, including positions excluded by the auxiliary gate. Appendix~\ref{app:auxiliary-selection} details the mask and loss normalization.

\begin{table*}[!t]
\centering
\footnotesize
\renewcommand{\arraystretch}{1.06}
\setlength{\tabcolsep}{2pt}
\begin{tabular*}{\textwidth}{@{\extracolsep{\fill}}cccccccc@{\hspace{7pt}}cccccc@{}}
\toprule
\multirow{2}{*}{\textbf{Benchmark}} & \multirow{2}{*}{\textbf{Metric}} & \multicolumn{6}{c}{\textbf{Qwen3-0.6B-Base (MATH)}} & \multicolumn{6}{c}{\textbf{Qwen3-1.7B-Base (MATH)}} \\
\cmidrule(lr){3-8}\cmidrule(lr){9-14}
 & & \textbf{Base} & \textbf{GRPO} & \textbf{OPD} & \textbf{TrOPD} & \textbf{EOPD} & \textbf{CCE-OPD} & \textbf{Base} & \textbf{GRPO} & \textbf{OPD} & \textbf{TrOPD} & \textbf{EOPD} & \textbf{CCE-OPD} \\
\midrule
\multirow{2}{*}{MATH500} & Avg@8 & 27.40 & 49.35 & 52.08 & 43.48 & \underline{52.45} & \cellcolor{cceblue}\textbf{54.00} & 53.78 & 63.58 & 68.45 & 59.98 & \underline{68.50} & \cellcolor{cceblue}\textbf{68.95} \\
 & Pass@8 & 67.60 & \textbf{77.60} & 77.00 & 73.20 & 76.80 & \cellcolor{cceblue}\underline{77.40} & 82.20 & 83.80 & \underline{86.40} & 86.00 & 85.40 & \cellcolor{cceblue}\textbf{87.60} \\
\midrule
\multirow{2}{*}{AIME24} & Avg@8 & 0.42 & 2.08 & \underline{3.75} & 2.50 & 3.33 & \cellcolor{cceblue}\textbf{4.58} & 3.33 & 6.25 & 9.17 & 9.17 & \underline{10.42} & \cellcolor{cceblue}\textbf{12.08} \\
 & Pass@8 & 3.33 & 6.67 & 6.67 & \textbf{16.67} & \underline{13.33} & \cellcolor{cceblue}10.00 & 16.67 & 20.00 & 20.00 & \textbf{26.67} & \underline{23.33} & \cellcolor{cceblue}\underline{23.33} \\
\midrule
\multirow{2}{*}{AIME25} & Avg@8 & 0.00 & \textbf{3.33} & \underline{2.08} & 0.83 & 1.67 & \cellcolor{cceblue}\textbf{3.33} & 1.25 & 4.17 & \underline{7.08} & 6.25 & 5.83 & \cellcolor{cceblue}\textbf{7.92} \\
 & Pass@8 & 0.00 & \textbf{20.00} & \underline{13.33} & 3.33 & \underline{13.33} & \cellcolor{cceblue}10.00 & 10.00 & \underline{16.67} & \underline{16.67} & \underline{16.67} & \underline{16.67} & \cellcolor{cceblue}\textbf{26.67} \\
\midrule
\multirow{2}{*}{AMC23} & Avg@8 & 9.69 & 22.19 & \underline{30.00} & 22.81 & 28.44 & \cellcolor{cceblue}\textbf{31.25} & 31.56 & 31.25 & 35.00 & 33.44 & \underline{35.31} & \cellcolor{cceblue}\textbf{40.00} \\
 & Pass@8 & 47.50 & 55.00 & 55.00 & \underline{60.00} & \underline{60.00} & \cellcolor{cceblue}\textbf{67.50} & 65.00 & \underline{67.50} & 62.50 & \underline{67.50} & 65.00 & \cellcolor{cceblue}\textbf{75.00} \\
\midrule
\multirow{2}{*}{Minerva} & Avg@8 & 5.19 & 12.22 & \underline{14.98} & 10.66 & 14.38 & \cellcolor{cceblue}\textbf{18.20} & 16.82 & \underline{25.18} & 24.08 & 22.01 & 21.51 & \cellcolor{cceblue}\textbf{26.01} \\
 & Pass@8 & 22.79 & 32.35 & \underline{34.93} & 29.78 & 31.62 & \cellcolor{cceblue}\textbf{35.66} & 39.34 & \textbf{45.59} & 42.65 & \textbf{45.59} & 38.24 & \cellcolor{cceblue}\underline{44.85} \\
\midrule
\multirow{2}{*}{Olympiad} & Avg@8 & 7.04 & 17.44 & \underline{20.35} & 17.28 & 20.13 & \cellcolor{cceblue}\textbf{21.50} & 20.98 & 27.63 & 31.83 & 30.24 & \underline{32.17} & \cellcolor{cceblue}\textbf{32.37} \\
 & Pass@8 & 27.70 & 40.30 & \underline{40.44} & 37.19 & 38.96 & \cellcolor{cceblue}\textbf{43.11} & 46.07 & 50.67 & 53.19 & \underline{54.07} & 53.33 & \cellcolor{cceblue}\textbf{54.22} \\
\bottomrule
\end{tabular*}
\caption{Mathematical reasoning results (\%). Bold and underlined values denote the best and second-best scores within each student scale and metric, including ties. Light blue highlights CCE-OPD. Evaluation protocols are specified in Section~\ref{sec:experimental-setup}.}
\label{tab:full}
\end{table*}

\section{Experiments}

\subsection{Experimental Setup}
\label{sec:experimental-setup}

\paragraph{Model and training data.}
We distill frozen Qwen3-8B into Qwen3-0.6B-Base and Qwen3-1.7B-Base \citep{yang2025qwen3} using 7,498 questions from the training split of the MATH dataset \citep{hendrycks2021math}. We compare Base, GRPO \citep{shao2024deepseekmath}, OPD \citep{li2026rethinking}, TrOPD \citep{xing2026trust}, and EOPD \citep{jin2026entropy}.

\paragraph{Evaluation.}
We sample eight responses per question at temperature 1.0 and top-$p=0.8$, with an 8,192-token limit. avg@8 is mean sampled-answer accuracy, and pass@8 is the fraction of questions solved at least once. Five mathematics tasks use the Qwen2.5-Math evaluator \citep{yang2024qwenmath}; OlympiadBench uses TINA/LightEval's \texttt{latex\_gold\_metric}. All methods share the task-specific evaluation protocol. Macro gains average the six differences between CCE-OPD and the best competing score for each benchmark. Dataset sizes appear in Appendix~\ref{app:training-details}.

\subsection{Main Results}

We evaluate whether CCE-OPD improves mathematical reasoning over standard OPD and other competing methods at both student scales. CCE-OPD achieves the highest avg@8 on all six 1.7B benchmarks and five 0.6B benchmarks, tying GRPO on 0.6B AIME25 (Table~\ref{tab:full}). At 1.7B, improvements over the strongest competing score range from 0.20 points on OlympiadBench to 4.69 on AMC23. At 0.6B, Minerva-Math shows the largest gain of 3.22 points. The corresponding macro gains are 1.45 and 1.33 points. CCE-OPD also obtains the best pass@8 on four tasks at 1.7B and three at 0.6B, showing improvements in both average answer accuracy and eight-attempt solution coverage.

\begin{table}[!t]
\centering
\footnotesize
\renewcommand{\arraystretch}{1.06}
\setlength{\tabcolsep}{2pt}
\begin{tabular*}{\columnwidth}{@{\extracolsep{\fill}}lccccc@{}}
\toprule
\multicolumn{6}{c}{Avg@8 (\%)} \\
\midrule
Benchmark & \cellcolor{cceblue}CCE-OPD & Random & Delete & Other & Symbols \\
\midrule
MATH500 & \cellcolor{cceblue}\textbf{68.95} & 67.38 & 68.23 & 67.03 & 67.58 \\
AIME24 & \cellcolor{cceblue}\textbf{12.08} & 11.25 & 9.58 & 9.17 & 9.17 \\
AIME25 & \cellcolor{cceblue}\textbf{7.92} & \textbf{7.92} & 4.58 & 5.83 & 5.83 \\
AMC23 & \cellcolor{cceblue}\textbf{40.00} & 37.81 & 39.69 & 38.75 & 39.69 \\
Minerva & \cellcolor{cceblue}\textbf{26.01} & 23.62 & 25.28 & 25.14 & 24.86 \\
Olympiad & \cellcolor{cceblue}\textbf{32.37} & 32.07 & 32.19 & 31.93 & 32.22 \\
\midrule
\multicolumn{6}{c}{Pass@8 (\%)} \\
\midrule
Benchmark & \cellcolor{cceblue}CCE-OPD & Random & Delete & Other & Symbols \\
\midrule
MATH500 & \cellcolor{cceblue}\textbf{87.60} & 84.60 & 86.00 & 85.40 & 86.00 \\
AIME24 & \cellcolor{cceblue}23.33 & 23.33 & 16.67 & 20.00 & \textbf{26.67} \\
AIME25 & \cellcolor{cceblue}\textbf{26.67} & 23.33 & 20.00 & 20.00 & 23.33 \\
AMC23 & \cellcolor{cceblue}\textbf{75.00} & \textbf{75.00} & \textbf{75.00} & 72.50 & 70.00 \\
Minerva & \cellcolor{cceblue}44.85 & 42.65 & 44.49 & 44.12 & \textbf{45.59} \\
Olympiad & \cellcolor{cceblue}54.22 & 53.48 & 53.63 & 54.52 & \textbf{54.81} \\
\bottomrule
\end{tabular*}
\caption{Second-input ablation for the 1.7B student ($\beta=0.2$). CCE-OPD masks the condition; Random masks a length-matched span; Delete removes the question; Other substitutes another question; Symbols matches its token count. Bold marks the best score in each row, including ties.}
\label{tab:question-ablation}
\end{table}

Compared with OPD, CCE-OPD improves avg@8 on all six benchmarks at both student scales, with macro-average gains of 1.60 and 1.95 percentage points for the 0.6B and 1.7B students, respectively. The largest gains occur on Minerva-Math for the 0.6B student (+3.22 points) and AMC23 for the 1.7B student (+5.00 points). These improvements span both student scales and benchmarks with different levels of baseline accuracy, supporting the use of condition contrast to improve the effectiveness of OPD across the evaluated mathematical reasoning tasks.

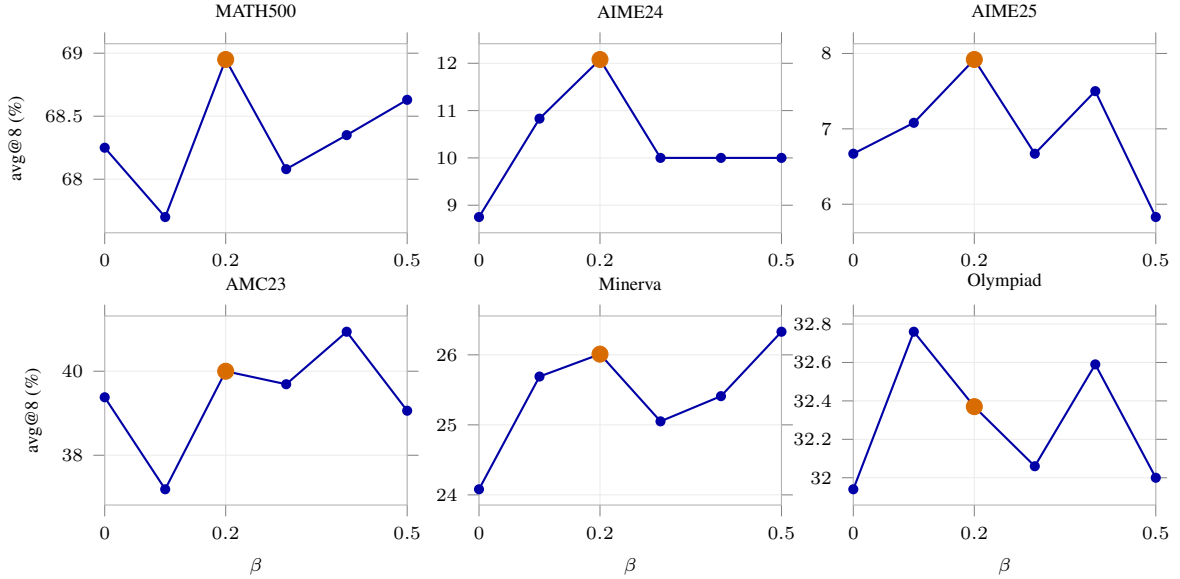
\begin{figure*}[!t]
\centering
\begin{tikzpicture}
\begin{groupplot}[group style={group size=3 by 2,horizontal sep=0.95cm,vertical sep=1.1cm},width=0.25\textwidth,height=2.5cm,scale only axis=true,title style={font=\scriptsize},tick label style={font=\scriptsize},label style={font=\scriptsize},xmin=0,xmax=0.5,xtick={0,0.2,0.5},xlabel={$\beta$},ylabel={avg@8 (\%)},grid=major,grid style={gray!15},axis line style={gray!60},tick align=outside,scaled ticks=false,/pgf/number format/fixed,/pgf/number format/precision=1]
\nextgroupplot[title={MATH500},xlabel={}]
\addplot[blue!65!black,thick,mark=*,mark size=1.5pt] coordinates {(0.0,68.25) (0.1,67.7) (0.2,68.95) (0.3,68.08) (0.4,68.35) (0.5,68.63)};
\addplot[orange!85!black,only marks,mark=*,mark size=3pt] coordinates {(0.2,68.95)};
\nextgroupplot[title={AIME24},xlabel={},ylabel={}]
\addplot[blue!65!black,thick,mark=*,mark size=1.5pt] coordinates {(0.0,8.75) (0.1,10.83) (0.2,12.08) (0.3,10) (0.4,10) (0.5,10)};
\addplot[orange!85!black,only marks,mark=*,mark size=3pt] coordinates {(0.2,12.08)};
\nextgroupplot[title={AIME25},xlabel={},ylabel={}]
\addplot[blue!65!black,thick,mark=*,mark size=1.5pt] coordinates {(0.0,6.67) (0.1,7.08) (0.2,7.92) (0.3,6.67) (0.4,7.5) (0.5,5.83)};
\addplot[orange!85!black,only marks,mark=*,mark size=3pt] coordinates {(0.2,7.92)};
\nextgroupplot[title={AMC23}]
\addplot[blue!65!black,thick,mark=*,mark size=1.5pt] coordinates {(0.0,39.38) (0.1,37.19) (0.2,40) (0.3,39.69) (0.4,40.94) (0.5,39.06)};
\addplot[orange!85!black,only marks,mark=*,mark size=3pt] coordinates {(0.2,40)};
\nextgroupplot[title={Minerva},ylabel={}]
\addplot[blue!65!black,thick,mark=*,mark size=1.5pt] coordinates {(0.0,24.08) (0.1,25.69) (0.2,26.01) (0.3,25.05) (0.4,25.41) (0.5,26.33)};
\addplot[orange!85!black,only marks,mark=*,mark size=3pt] coordinates {(0.2,26.01)};
\nextgroupplot[title={Olympiad},ylabel={}]
\addplot[blue!65!black,thick,mark=*,mark size=1.5pt] coordinates {(0.0,31.94) (0.1,32.76) (0.2,32.37) (0.3,32.06) (0.4,32.59) (0.5,32)};
\addplot[orange!85!black,only marks,mark=*,mark size=3pt] coordinates {(0.2,32.37)};
\end{groupplot}
\end{tikzpicture}
\caption{Sensitivity to extrapolation strength on the six mathematics benchmarks (1.7B student). Orange marks the main setting $\beta=0.2$. Each panel uses its own vertical scale. Full avg@8 and pass@8 scores appear in Table~\ref{tab:beta}.}
\label{fig:beta}
\end{figure*}

\subsection{Which Part of the Question Should Be Masked?}

To assess whether the benefit depends on masking a condition, we vary the second teacher input on the same 3,402 eligible questions at 1.7B and $\beta=0.2$. Alternatives are a length-matched random span that preserves the request, whole-question deletion retaining format instructions, another question from a fixed training-set permutation, and symbols matched to the question's token count. Each arm retains the original questions for student rollouts and the same loss settings.

These controls examine distinct aspects of the second teacher view. Random tests the value of selecting a specific condition under matched masking lengths. Delete tests the role of retaining the remaining question information. Other tests whether replacing the question with different mathematical content yields similar benefits, while Symbols tests whether preserving input length without meaningful question content is sufficient. Together, these comparisons assess the contribution of condition masking.

Condition masking improves avg@8 over random masking on five benchmarks and ties on AIME25 (Table~\ref{tab:question-ablation}). Its largest advantages are 2.39 points on Minerva-Math and 2.19 on AMC23. It also outperforms each broader question perturbation on all six avg@8 metrics. The comparison links the benefit of the second teacher view to the condition selected for masking. Changes to pass@8 are less uniform, particularly for Minerva-Math and OlympiadBench.

\subsection{How Much Extrapolation Is Useful?}
\label{sec:beta-ablation}

Having examined the input construction, we next evaluate sensitivity to extrapolation strength by varying $\beta\in\{0,0.1,0.2,0.3,0.4,0.5\}$ for the 1.7B student. The $\beta=0$ arm retains the paired teacher queries, candidate union, entropy gate, and auxiliary loss, allowing the effect of reweighting to be measured within the same training setup.

Relative to $\beta=0$, the main setting $\beta=0.2$ improves avg@8 on all six tasks, by 0.70, 3.33, 1.25, 0.62, 1.93, and 0.43 points in benchmark order (Figure~\ref{fig:beta}). It improves pass@8 on five tasks and ties on Minerva-Math. Larger coefficients do not yield monotonic gains: $\beta=0.4$ leads on AMC23, while $\beta=0.5$ leads on Minerva-Math. The results favor a moderate adjustment to the full teacher, with task-dependent sensitivity to its strength.

\subsection{Does Entropy Gating Improve Distillation?}
\label{sec:entropy-ablation}

To assess the value of selective auxiliary supervision, we compare the entropy gate with applying the auxiliary loss to every valid response position on eligible questions. Both arms use $\beta=0.2$, $\lambda=1$, and normalization by all valid response tokens. The remaining 4,096 questions receive OPD in both arms.

\begin{table}[!t]
\centering
\footnotesize
\renewcommand{\arraystretch}{1.08}
\setlength{\tabcolsep}{3pt}
\begin{tabular*}{\columnwidth}{@{\extracolsep{\fill}}lcccc@{}}
\toprule
& \multicolumn{2}{c}{Entropy gate} & \multicolumn{2}{c}{No gate} \\
\cmidrule(lr){2-3}\cmidrule(lr){4-5}
Benchmark & Avg@8 & Pass@8 & Avg@8 & Pass@8 \\
\midrule
MATH500 & \cellcolor{cceblue}\textbf{68.95} & \cellcolor{cceblue}\textbf{87.60} & 68.65 & 87.20 \\
AIME24 & \cellcolor{cceblue}\textbf{12.08} & \cellcolor{cceblue}\textbf{23.33} & 10.83 & \textbf{23.33} \\
AIME25 & \cellcolor{cceblue}\textbf{7.92} & \cellcolor{cceblue}\textbf{26.67} & 6.67 & 23.33 \\
AMC23 & \cellcolor{cceblue}\textbf{40.00} & \cellcolor{cceblue}\textbf{75.00} & 39.38 & 72.50 \\
Minerva & \cellcolor{cceblue}\textbf{26.01} & \cellcolor{cceblue}44.85 & 23.53 & \textbf{45.59} \\
Olympiad & \cellcolor{cceblue}\textbf{32.37} & \cellcolor{cceblue}\textbf{54.22} & 31.44 & 53.93 \\
\bottomrule
\end{tabular*}
\caption{Entropy-gate ablation for the 1.7B student ($\beta=0.2$; scores in \%). The gate selects positions with $H(q_T)\geq0.8$ nats; No gate supervises all valid response positions on eligible questions. Bold marks the best score for each benchmark and metric, including ties.}
\label{tab:entropy-gate}
\end{table}

The gated configuration improves avg@8 on all six tasks (Table~\ref{tab:entropy-gate}), with gains ranging from 0.30 points on MATH500 to 2.48 on Minerva-Math. It also improves pass@8 on four tasks and ties on AIME24. Concentrating the auxiliary loss at uncertain teacher positions is therefore effective under the fixed coefficient and normalization used here.

The macro-average avg@8 gain is 1.14 percentage points, with improvements of 1.25 points on both AIME24 and AIME25. Minerva-Math illustrates the distinction between the two metrics: gating raises avg@8 by 2.48 points while reducing pass@8 by 0.74 points, indicating higher average answer accuracy alongside slightly lower eight-attempt coverage. Because the loss coefficient and denominator are fixed, removing the gate adds auxiliary contributions from low-entropy positions. Their inclusion does not improve avg@8 on any evaluated benchmark. This pattern supports the selection rule in Section~\ref{sec:cce-method}: focusing condition-based preference refinement on positions where the teacher distributes probability across competing continuations.

\subsection{Transfer Beyond Mathematics}

Finally, to test transfer beyond the training domain, we evaluate the mathematics-trained 1.7B models on GPQA-Diamond \citep{rein2023gpqa}, MMLU-Pro \citep{wang2024mmlupro}, IFEval \citep{zhou2023ifeval}, and CRUXEval-O \citep{gu2024cruxeval}, using a shared task-specific protocol across models.

\begin{table}[!t]
\centering
\footnotesize
\renewcommand{\arraystretch}{1.2}
\setlength{\tabcolsep}{5pt}
\begin{tabular}{@{}lcccc@{}}
\toprule
Method & \shortstack{GPQA\\ D} & \shortstack{MMLU\\ Pro} & IFEval & \shortstack{CRUX\\ O} \\
\midrule
Base & 24.24 & 35.90 & 16.64 & 34.2250 \\
OPD & 29.29 & 40.23 & 29.57 & 39.5125 \\
EOPD & 29.29 & 39.99 & 30.50 & 40.3625 \\
\addlinespace[2pt]
\rowcolor{cceblue} \textbf{CCE-OPD} & \textbf{30.81} & \textbf{40.72} & \textbf{32.53} & \textbf{41.1125} \\
\bottomrule
\end{tabular}
\caption{Cross-domain results for 1.7B students (\%). Bold denotes the best score per task. CRUX-O denotes CRUXEval-O.}
\label{tab:cross-domain}
\end{table}

CCE-OPD obtains the highest score on all four transfer tasks (Table~\ref{tab:cross-domain}). Gains over the strongest competing result are 1.52 points on GPQA-Diamond, 0.49 on MMLU-Pro, 2.03 on IFEval, and 0.75 on CRUXEval-O. These results extend the observed benefits to knowledge, instruction following, and code reasoning after mathematics-only training.

\section{Related Work}

\paragraph{On-policy distillation.}
OPD extends sequence-level distillation \citep{kim2016sequence} by learning on student-generated trajectories \citep{agarwal2023gkd,gu2024minillm}, connecting to earlier work on learner-generated histories \citep{ross2011dagger,bengio2015scheduled}. DistiLLM improves loss design and sample reuse \citep{ko2024distillm}. Further studies examine supervision failures \citep{li2026rethinking,fu2026revisiting,liu2026fidelity}, stabilize targets or select compatible positions \citep{jang2026stable,xing2026trust,wang2026teachability}, and provide step-level corrections \citep{sun2026step}. EOPD adds forward KL at high-entropy teacher positions \citep{jin2026entropy}, while ExOPD scales the distillation reward relative to KL regularization \citep{yang2026exopd}. CCE-OPD also uses entropy-gated auxiliary supervision, refining its target through condition contrast to reduce preferences for CVC.

\paragraph{Contrastive and extrapolative supervision.}
Predictive contrasts guide generation across models \citep{liu2021dexperts,li2022contrastive}, control codes \citep{krause2021gedi}, layers \citep{chuang2024dola}, and context views \citep{shi2023cad}. Proxy-tuning transfers the difference between tuned and untuned small-model predictions to a larger model \citep{liu2024proxy}. For distillation, DistiLLM-2 contrasts teacher- and student-response objectives \citep{ko2025distillm2}, while Delta Distillation and weak-to-strong OPD construct supervision from capability contrasts \citep{heo2026delta,yu2026weak}. CCE-OPD shares the unclipped probability-ratio form of context-aware decoding. It compares full and masked questions under the same frozen teacher and student prefix, using bounded extrapolation to form an auxiliary training target.

\section{Conclusion}

Teacher feedback on student trajectories can assign positive local advantage to a sampled token that begins a CVC. CCE-OPD uses the teacher’s response to a masked condition to refine that supervision. The paired diagnostic shows improved candidate ranking, and training experiments show higher or tied-highest avg@8 on six mathematical benchmarks at two student scales, together with gains on four transfer tasks. The input-construction ablations highlight the value of selecting a relevant condition for masking. The extrapolation-strength and entropy-gate ablations further support moderate reweighting at positions with competing teacher predictions. The results support condition contrast as a practical way to refine continuation preferences in on-policy distillation.

\section*{Limitations}

Our training experiments use mathematical reasoning data from the MATH dataset. Future work could extend CCE-OPD to training data from a broader range of domains, such as code generation and scientific reasoning, to investigate its effectiveness across different types of tasks.

\section*{Ethical Considerations}

We use existing benchmark datasets and pretrained Qwen models, whose use and redistribution are governed by the applicable licenses and access conditions. Distilled students may inherit biases or undesirable behaviors from their teachers, warranting evaluation before downstream deployment.

DeepSeek-Flash assists condition annotation and semantic review, while ChatGPT-6 Sol assists mathematical verification of diagnostic candidates. The authors are responsible for the annotation decisions and reported results.

\bibliography{custom}

\appendix
\section{Mathematical Verification of the Local Example}
\label{app:error-context}

The question is: ``You are given that $3^{400}\equiv1\pmod{1000}$. What are the last three digits of $3^{12000}$?'' The given congruence yields $3^{12000}=(3^{400})^{30}\equiv1\pmod{1000}$, so the requested digits are 001. It does not identify the smallest positive exponent giving residue 1.

Direct modular calculations give
\begin{align*}
3^{100}&\equiv1\pmod{1000},\\
3^{50}&\equiv249\pmod{1000},\\
3^{20}&\equiv401\pmod{1000}.
\end{align*}
The multiplicative order divides 100. Every proper divisor of 100 divides either 50 or 20, ruling out all smaller possibilities. The order is therefore exactly 100.

Before the erroneous assertion, the student has restated the given congruence. The scored prefix ends immediately before the first digit in ``the order of 3 modulo 1000 is **400**''. The two decision tokens begin the complete assertions with values 400 and 100. Their probabilities describe this specific decision, while other tokens can begin valid continuations or CVC. In the masked question, the given-congruence clause is replaced by \texttt{[MASKED]}; the student prefix is retained verbatim.

\section{Candidate-Preference Diagnostic}
\label{app:paired-context}

\paragraph{State construction.}
To examine continuation preferences at student reasoning states, we construct the diagnostic set from the 3,402 questions eligible for condition masking in the MATH training split. An OPD-trained Qwen3-1.7B student generates one response per question with temperature 1.0, top-$p=1.0$, thinking disabled, and a maximum response length of 8,096 tokens. Questions are processed in a predetermined hash-based order, excluding previously examined pilot questions. We retain the first 200 qualifying states, with at most one state per question.

Next, for each response, we identify the first substantive inference that uses the selected condition and truncate the student prefix immediately before that inference. States with an earlier uncorrected mathematical error or an ambiguous candidate pair are excluded. The original inference forms one candidate; a minimally edited alternative forms the other, yielding a valid continuation and a CVC with comparable wording and the same reasoning purpose. Candidate validity is established through ChatGPT-6 Sol-assisted mathematical verification against the question and prefix, before teacher candidate scores are inspected. The resulting set contains 200 states, including 60 whose naturally generated next inference is erroneous.

\paragraph{Scoring protocol.}
With the states and candidate pairs fixed, the frozen Qwen3-8B teacher evaluates identical prefixes and candidate sequences under the complete question, the masked question, and a question with length-matched random masking. For a prefix $h$ and a candidate inference $z=(z_1,\ldots,z_L)$, we use the mean token log-probability score
\begin{equation}
S_q(z)=\frac{1}{L}\sum_{j=1}^{L}\log q(z_j\mid x,h,z_{<j}).
\end{equation}
Each candidate supplies its own continuation history during teacher forcing, and the teacher under the masked question substitutes $x^{-}$ for $x$. We then score each candidate using the normalized condition-contrastive target:
\begin{equation}
S_{\mathrm{contrast}}(z)=S_T(z)+\frac{1}{L}\sum_{j=1}^{L}\bigl[\beta\tilde\delta_j(z_j)-\log Z_j\bigr],
\end{equation}
where $\tilde\delta_j$ is the condition contrast clipped to $[-5,5]$, $\beta=0.2$, and $Z_j$ is the per-position normalizing constant of $q_E$ at the candidate-specific history $(x,h,z_{<j})$. The normalizer is evaluated separately at each candidate's own history before sequence averaging. Thus, $S_{\mathrm{contrast}}(z)$ is the mean token log probability under the normalized target, following the scoring procedure in Section~\ref{sec:paired-diagnostic}.

For a controlled comparison, random masking preserves the requested output and response-format instruction and matches the number of source tokens masked by the condition-masking intervention. We average the normalized contrast scores over three fixed random-masking views before ranking the candidates. All question variants retain the same student prefix, including any earlier restatement of the selected condition. The reported preference rate counts states in which the CVC receives a higher score than the valid candidate. This span-level diagnostic complements the single-token analysis in Section~\ref{sec:error-signal}.

\paragraph{Paired preference changes.}
Finally, we compare the full-teacher and condition-augmented rankings state by state. Condition contrast corrects 23 full-teacher preferences and reverses one previously correct preference, reducing the number of states preferring the CVC from 54 to 32. All 24 reversals occur within the 60-state error subset. Preferences remain unchanged in the remaining 140 states, five of which favor the CVC under both scoring schemes. Table~\ref{tab:paired-preference-changes} summarizes these transitions.

\begin{table}[H]
\centering
\small
\setlength{\tabcolsep}{4pt}
\begin{tabular*}{\columnwidth}{@{\extracolsep{\fill}}lrr@{}}
\toprule
Preference change & All states (200) & Error subset (60) \\
\midrule
CVC $\rightarrow$ valid & 23 & 23 \\
Valid $\rightarrow$ CVC & 1 & 1 \\
Unchanged & 176 & 36 \\
\bottomrule
\end{tabular*}
\caption{Paired preference changes after adding condition contrast to the full-teacher score.}
\label{tab:paired-preference-changes}
\end{table}

Together, these paired comparisons show how condition contrast changes continuation rankings at fixed student states. The reported proportions describe preferences within the diagnostic set, rather than error rates during unrestricted generation.


\begin{table*}[t]
\centering
\footnotesize
\renewcommand{\arraystretch}{1.12}
\setlength{\tabcolsep}{5pt}
\begin{tabular}{@{}p{0.32\textwidth}p{0.32\textwidth}p{0.31\textwidth}@{}}
\toprule
Original question & Masked question & Rationale \\
\midrule
\textit{Numerical information.} Margo walks to her friend's house in 10 minutes. Using the same route, it takes Margo 20 minutes to get back home. \textbf{If her average walking rate for the entire trip is 4 miles per hour}, how many total miles did she walk?
& Margo walks to her friend's house in 10 minutes. Using the same route, it takes Margo 20 minutes to get back home. \texttt{[MASKED]}, how many total miles did she walk?
& The average speed connects the total travel time to the requested distance, giving $4\times(30/60)=2$ miles. Masking it removes the information needed to determine that distance. \\
\midrule
\textit{Geometric relation.} A rectangular prism has dimensions 8 inches by 2 inches by 32 inches. \textbf{If a cube has the same volume as the prism}, what is the surface area of the cube, in square inches?
& A rectangular prism has dimensions 8 inches by 2 inches by 32 inches. \texttt{[MASKED]}, what is the surface area of the cube, in square inches?
& The equal-volume relation determines the cube's side length as $\sqrt[3]{8\times2\times32}=8$ and its surface area as $6\times8^2=384$. Masking it leaves the cube's size unspecified. \\
\midrule
\textit{Complex-variable constraint.} Let $z$ be a nonreal complex number \textbf{such that $|z|=1$}. Find the real part of $\frac{1}{1-z}$.
& Let $z$ be a nonreal complex number \texttt{[MASKED]}. Find the real part of $\frac{1}{1-z}$.
& The unit-modulus constraint fixes the requested real part at $1/2$. Without it, the real part varies with $z$: it equals $1/2$ for $z=i$ and $1/5$ for $z=2i$. \\
\bottomrule
\end{tabular}
\caption{Masked inputs from the MATH training set. Bold identifies the selected source span; \texttt{[MASKED]} denotes the masking marker. The shared response-format instruction is omitted.}
\label{tab:condition-construction-examples}
\end{table*}

\section{Auxiliary-Loss Position Selection}
\label{app:auxiliary-selection}

Following Section~\ref{sec:cce-method}, auxiliary supervision is applied only when the question has an accepted masked counterpart, the position is a valid response token, and the full-input teacher entropy satisfies $H(q_T)\geq0.8$ nats. Entropy is computed over the complete vocabulary using natural logarithms, before candidate-support construction and renormalization. The same mask is used in the main experiments and the input-construction and extrapolation-strength ablations; the no-gate ablation removes only the entropy criterion. In all settings, the auxiliary loss is normalized by the total number of valid response tokens in the update.

\section{Construction and Training Details}
\label{app:training-details}

\paragraph{Condition construction.}
The annotator proposes one minimal, semantically complete condition and the smallest span specifying the requested output. Condition types include quantities, constraints, relations, domain restrictions, structural properties, and definitions. The two proposed spans must match the source exactly and must not overlap. Semantic review checks that the masked condition affects an answer constraint or necessary solution path while the request remains identifiable. Accepted spans are replaced with a fixed masking marker before training. For presentation, we denote this marker by \texttt{[MASKED]}, following Figure~\ref{fig:cce-method}.

To illustrate the resulting inputs, Table~\ref{tab:condition-construction-examples} presents three examples from the MATH training set. They cover numerical information, a geometric relation, and a constraint on a complex variable. In each case, masking the condition changes the information available for solving the question while retaining the requested output.

\paragraph{Length-matched random masking.}
For comparison, Table~\ref{tab:random-cover-example} shows a saved random-masking view of the geometric example above. Both source spans contain 10 tokens under the shared Qwen tokenizer and are replaced with the same masking marker. The random window is sampled from the question while preserving the requested output; such windows may mask other relevant information or overlap the selected condition.

\begin{table}[H]
\centering
\footnotesize
\setlength{\tabcolsep}{4pt}
\begin{tabular}{@{}p{0.19\columnwidth}p{0.59\columnwidth}r@{}}
\toprule
Masking & Source span & Tokens \\
\midrule
Condition & If a cube has the same volume as the prism & 10 \\
Random span & dimensions 8 inches by 2 inches by & 10 \\
\bottomrule
\end{tabular}
\caption{An example of token-length matching between condition masking and random masking. Lengths refer to the replaced source spans.}
\label{tab:random-cover-example}
\end{table}

The resulting randomly masked question is:
\begin{quote}
\small
A rectangular prism has\texttt{[MASKED]}32 inches. If a cube has the same volume as the prism, what is the surface area of the cube, in square inches?
\end{quote}


\paragraph{Annotation prompts.}
We use DeepSeek-Flash with thinking disabled for both candidate annotation and semantic review. The system message for candidate annotation is: \texttt{You are a careful mathematical dataset annotator.} The user template below is instantiated with the original question in the final \texttt{Problem} field.

\begin{lstlisting}[style=cceprompt]
For the Problem below, identify the smallest target text that must remain and exactly one smallest semantically complete premise whose removal materially changes the constraints, answer determinacy, or necessary solution path while leaving a coherent question.

Return exactly one JSON object and no explanation.

Accepted schema:
{"status":"accepted","target":{"text":"exact source substring"},"spans":[{"text":"exact source substring","kind":"constraint","provenance":"short explanation","confidence":0.0}]}

Rejected schema:
{"status":"rejected","rejection_reason":"short reason","target":null,"spans":[]}

Hard rules:
- Every text value in an accepted response must be copied from Problem as one contiguous, character-for-character substring.
- Never paraphrase, normalize, correct, translate, reformat LaTeX, alter backslashes, alter punctuation, or alter whitespace in copied text.
- The target and removed condition must not overlap.
- Select exactly one condition. It must be an atomic, semantically complete proposition, not a fragment such as only a number, a conjunction, or a trailing phrase.
- The target must still identify what answer is requested after the condition is removed.
- kind must be one of: quantity, constraint, relation, domain, structural_property, definition.
- Reject rather than guess when no separable target and removable premise exist.

Problem:
{problem}
\end{lstlisting}

\paragraph{Validation and semantic review.}
For an accepted response, the validator resolves the target and condition to unique, nonempty source substrings and records their character offsets. It checks that exactly one condition is returned, the two spans do not overlap, the condition kind belongs to the specified set, the provenance is nonempty, and the confidence lies in $[0,1]$. Candidates passing these checks are submitted to a second DeepSeek-Flash call using the following template, with the question and extracted spans inserted into the corresponding fields.

\begin{lstlisting}[style=cceprompt]
Judge one proposed CCE premise deletion. Return only JSON:
{"decision":"accept" or "reject","reason":"max 12 words"}

Accept when Condition is one semantically complete mathematical premise, definition, domain restriction, or geometric construction; it must materially constrain the answer or a necessary solution path, and Target must remain the requested output. It is VALID and desired if deleting Condition makes the remaining question underdetermined or unsolvable.

Reject only output-format instructions, text fragments, the requested expression/question itself rather than a premise, duplicated/leaked information, or information immaterial to solving. Do NOT reject merely because Condition supplies all decisive constraints or is a necessary definition/construction.

Problem: {problem}
Target: {target}
Condition: {condition}
\end{lstlisting}

The semantic review retains conditions that affect the answer constraints or a necessary solution path while preserving the requested output. A question may become underdetermined after masking; the retained request must still identify the quantity or expression to be answered. The accepted condition spans are frozen before training, and the response-format instruction is preserved in both question views. This process yields the 3,402 eligible questions used for auxiliary supervision.

\paragraph{Optimization.}
Table~\ref{tab:hyperparameters} summarizes the optimization, rollout, and evaluation settings used in our experiments, together with the main CCE-OPD hyperparameters.

\begin{table}[!t]
\centering
\footnotesize
\renewcommand{\arraystretch}{1.05}
\setlength{\tabcolsep}{3pt}
\begin{tabular*}{\columnwidth}{@{\extracolsep{\fill}}lr@{}}
\toprule
Hyperparameter & Value \\
\midrule
\multicolumn{2}{l}{\textit{Optimization}} \\
Optimizer & AdamW \\
Learning rate & $3\times10^{-6}$ \\
Learning-rate schedule & Cosine \\
Weight decay & 0.01 \\
Gradient clipping norm & 1.0 \\
PPO epochs per update & 1 \\
\midrule
\multicolumn{2}{l}{\textit{Training rollouts}} \\
Maximum prompt length & 2,048 tokens \\
Maximum response length & 2,048 tokens \\
Sampling temperature & 1.0 \\
Top-$p$ & 1.0 \\
Thinking mode & Disabled \\
\midrule
\multicolumn{2}{l}{\textit{CCE-OPD}} \\
Top-$k$ per teacher distribution & 16 \\
Extrapolation coefficient $\beta$ & 0.2 \\
Contrast clipping bound $c$ & 5 \\
Entropy threshold $\tau$ & 0.8 nats \\
Auxiliary-loss coefficient $\lambda$ & 1 \\
\midrule
\multicolumn{2}{l}{\textit{Evaluation}} \\
Responses per question & 8 \\
Sampling temperature & 1.0 \\
Top-$p$ & 0.8 \\
Maximum response length & 8,192 tokens \\
\bottomrule
\end{tabular*}
\caption{Hyperparameter settings for training and evaluation. CCE-OPD uses the same main settings for both student scales; the extrapolation-strength and entropy-gate ablations vary their respective settings.}
\label{tab:hyperparameters}
\end{table}

\paragraph{Evaluation sets.}
We evaluate MATH500, a subset of MATH \citep{hendrycks2021math}; AIME 2024; AIME 2025; AMC 2023; Minerva-Math \citep{lewkowycz2022minerva}; and OlympiadBench \citep{he2024olympiadbench}. Their sizes are 500, 30, 30, 40, 272, and 675 questions, respectively. Eight responses per question yield 12,376 evaluated responses per checkpoint. Benchmark scores use the task evaluators described in Section~\ref{sec:experimental-setup}; training rewards are not used to populate the evaluation tables.

\section{Full Extrapolation-Strength Results}
\label{app:beta-results}
Table~\ref{tab:beta} reports the complete scores corresponding to Figure~\ref{fig:beta}.

\begin{table}[!t]
\centering
\footnotesize
\renewcommand{\arraystretch}{1.06}
\setlength{\tabcolsep}{2pt}
\begin{tabular*}{\columnwidth}{@{\extracolsep{\fill}}lcccccc@{}}
\toprule
\multicolumn{7}{c}{Avg@8 (\%)} \\
\midrule
Benchmark & $0$ & $0.1$ & \cellcolor{cceblue}$0.2$ & $0.3$ & $0.4$ & $0.5$ \\
\midrule
MATH500 & 68.25 & 67.70 & \cellcolor{cceblue}\textbf{68.95} & 68.08 & 68.35 & 68.63 \\
AIME24 & 8.75 & 10.83 & \cellcolor{cceblue}\textbf{12.08} & 10.00 & 10.00 & 10.00 \\
AIME25 & 6.67 & 7.08 & \cellcolor{cceblue}\textbf{7.92} & 6.67 & 7.50 & 5.83 \\
AMC23 & 39.38 & 37.19 & \cellcolor{cceblue}40.00 & 39.69 & \textbf{40.94} & 39.06 \\
Minerva & 24.08 & 25.69 & \cellcolor{cceblue}26.01 & 25.05 & 25.41 & \textbf{26.33} \\
Olympiad & 31.94 & \textbf{32.76} & \cellcolor{cceblue}32.37 & 32.06 & 32.59 & 32.00 \\
\midrule
\multicolumn{7}{c}{Pass@8 (\%)} \\
\midrule
Benchmark & $0$ & $0.1$ & \cellcolor{cceblue}$0.2$ & $0.3$ & $0.4$ & $0.5$ \\
\midrule
MATH500 & 87.20 & 86.20 & \cellcolor{cceblue}\textbf{87.60} & 87.40 & 87.00 & 86.80 \\
AIME24 & 16.67 & \textbf{30.00} & \cellcolor{cceblue}23.33 & 20.00 & 20.00 & \textbf{30.00} \\
AIME25 & 20.00 & 20.00 & \cellcolor{cceblue}\textbf{26.67} & 16.67 & 20.00 & 20.00 \\
AMC23 & 70.00 & 62.50 & \cellcolor{cceblue}\textbf{75.00} & 72.50 & 67.50 & 70.00 \\
Minerva & 44.85 & 44.85 & \cellcolor{cceblue}44.85 & 41.91 & 45.96 & \textbf{46.69} \\
Olympiad & 53.93 & \textbf{55.70} & \cellcolor{cceblue}54.22 & 55.11 & 55.26 & 54.37 \\
\bottomrule
\end{tabular*}
\caption{Extrapolation strength $\beta$ for the 1.7B student. The $\beta=0$ control retains auxiliary supervision without condition reweighting. Blue marks the main setting; bold marks the best score in each row, including ties.}
\label{tab:beta}
\end{table}

\section{CCE-OPD Training Algorithm}
\label{app:cce-algorithm}

The complete question is used for every rollout. For eligible questions, both teacher evaluations retain the same student-generated prefix. Teacher predictions, rollout probabilities, and the auxiliary target are held fixed during the update; gradients flow through the current student probabilities. We use $k=16$, $\beta=0.2$, $c=5$, $\tau=0.8$ nats, and $\lambda=1$.

\begin{algorithm}[!t]
\caption{CCE-OPD training}
\label{alg:cce-opd}
\footnotesize
\begin{algorithmic}[1]
\Require Student $p_\theta$, frozen teacher $q$, questions $\mathcal D$, and accepted counterparts $x^{-}$
\Require $k$, $\beta$, $c$, entropy threshold $\tau$, and coefficient $\lambda$
\Repeat
  \State Freeze the rollout policy $p_{\mathrm{old}}\gets p_\theta$
  \State Sample questions from $\mathcal D$ and responses from $p_{\mathrm{old}}(\cdot\mid x)$
  \State Store sampled-token rollout probabilities
  \State Score all valid prefixes with $q_T=q(\cdot\mid x,\hat y_{<t})$
  \State Score eligible prefixes with $q_D=q(\cdot\mid x^{-},\hat y_{<t})$
  \State $A_t\gets\log q_T(\hat y_t)-\log p_{\mathrm{old}}(\hat y_t)$
  \State Compute current $p_{\theta,t}$ over the full vocabulary
  \State $N\gets$ total number of valid response tokens
  \State $L_{\mathrm{aux}}\gets0$
  \For{each eligible valid response position $t$}
    \State $H_t\gets-\sum_{v\in\mathcal V}q_T(v)\log q_T(v)$
    \If{$H_t\geq\tau$}
      \State $\mathcal V_t\gets\operatorname{TopK}(q_T,k)\cup\operatorname{TopK}(q_D,k)$
      \State $\delta_t(v)\gets\log q_T(v)-\log q_D(v)$, $v\in\mathcal V_t$
      \State $\tilde\delta_t(v)\gets\operatorname{clip}(\delta_t(v),-c,c)$
      \State $w_t(v)\gets\log q_T(v)+\beta\tilde\delta_t(v)$
      \State $q_E\gets\operatorname{softmax}_{\mathcal V_t}(w_t)$
      \State $L_{\mathrm{aux}}\gets L_{\mathrm{aux}}+D_{\mathrm{KL}}(q_E\|p_{\theta,t})$
    \EndIf
  \EndFor
  \State Compute $\mathcal L_{\mathrm{OPD}}$ using $A_t$ and the clipped policy ratios over all valid response tokens
  \State $\mathcal L_{\mathrm{CCE}}\gets L_{\mathrm{aux}}/N$
  \State Update $\theta$ using $\mathcal L_{\mathrm{OPD}}+\lambda\mathcal L_{\mathrm{CCE}}$
\Until{training is complete}
\State \Return $p_\theta$
\end{algorithmic}
\end{algorithm}

The candidate union is deduplicated and contains at most $2k$ tokens. Condition contrast uses teacher log probabilities normalized over the complete vocabulary; only $q_E$ is normalized on $\mathcal V_t$. The student probabilities in the auxiliary KL retain full-vocabulary normalization, and $q_E$ is zero outside its support. Positions excluded by eligibility or entropy selection contribute zero auxiliary loss while retaining sampled-token OPD supervision. The denominator $N$ includes all valid response tokens, as in Section~\ref{sec:cce-method}.

\section{Transfer Evaluation Protocols}
\label{app:transfer-protocols}
We evaluate the mathematics-trained 1.7B models on four transfer benchmarks using the same task-specific prompts, decoding settings, and scoring procedures across methods. Thinking mode is disabled throughout. Table~\ref{tab:transfer-protocols} summarizes the evaluation settings.

\begin{table}[!t]
\centering
\footnotesize
\renewcommand{\arraystretch}{1.10}
\setlength{\tabcolsep}{2pt}
\begin{tabular*}{\columnwidth}{@{\extracolsep{\fill}}lrrrr@{}}
\toprule
Benchmark & Questions & Shots & \shortstack{Max.\\tokens} & \shortstack{Responses\\per question} \\
\midrule
GPQA-Diamond & 198 & 0 & -- & -- \\
MMLU-Pro & 12,032 & 5 & 2,048 & 1 \\
IFEval & 541 & 0 & 1,280 & 1 \\
CRUXEval-O & 800 & 2 & 8,096 & 10 \\
\bottomrule
\end{tabular*}
\caption{Transfer evaluation settings. GPQA-Diamond scores four options by log likelihood. MMLU-Pro and IFEval use greedy decoding; CRUXEval-O uses temperature 0.2 and top-$p=0.95$.}
\label{tab:transfer-protocols}
\end{table}

\paragraph{Prompts.}
GPQA-Diamond, IFEval, and CRUXEval-O use the Qwen tokenizer's chat template. GPQA-Diamond presents each question with four answer options and selects the option with the highest conditional log likelihood. MMLU-Pro uses a subject-specific multiple-choice prompt with five demonstrations in plain text; generation stops at \texttt{Question:} or the token limit. IFEval directly supplies the benchmark instruction as the user prompt. CRUXEval-O provides a Python function and an incomplete assertion, \texttt{assert f(input) == ??}, and requests the literal output in a complete assertion enclosed in \texttt{[ANSWER]} and \texttt{[/ANSWER]} tags. Its two fixed demonstrations are formatted as preceding user--assistant turns.

\paragraph{Scoring and versions.}
GPQA-Diamond and IFEval use task versions \texttt{2.2-local-56686c06} and \texttt{4.0-local-966cd895}, respectively. GPQA-Diamond reports option accuracy. IFEval reports strict prompt-level accuracy, requiring every instruction associated with a prompt to be satisfied. MMLU-Pro uses task version \texttt{3.1-local-b189ec76} and scorer \texttt{mmlu-pro-saved-final-choice-v1}, which extracts a terminal answer letter or boxed choice, falls back to the original \texttt{answer is} filter, and compares the extracted choice with the gold answer to report exact-match accuracy.

CRUXEval-O uses task version \texttt{1.1-local-crux8096-20260925}. Its scoring pipeline comprises \texttt{cruxeval-saved-cpu-v1}, \texttt{crux8096-final-assertion-\allowbreak diagnostic-v1}, and \texttt{crux8096-output-final-assert-\allowbreak input-checked-v1}. The pipeline extracts the final assertion, verifies that its function-call input matches the benchmark input, and checks correctness by execution with a three-second timeout. The reported pass@1 averages the ten-response success fraction over all 800 questions (8,000 responses).

\end{document}